\documentclass[conference]{IEEEtran}
\IEEEoverridecommandlockouts
\usepackage{cite}
\usepackage{amsmath,amssymb,amsfonts}
\usepackage{algorithmic}
\usepackage{graphicx}
\usepackage{textcomp}
\usepackage{xcolor}
\usepackage{booktabs}
\usepackage{array}
\usepackage{makecell}
\usepackage{tabularx}
\usepackage{multirow}
\usepackage{amsmath}
\def\BibTeX{{\rm B\kern-.05em{\sc i\kern-.025em b}\kern-.08em
    T\kern-.1667em\lower.7ex\hbox{E}\kern-.125emX}}
\begin{document}

\title{PCM-SAR: Physics-Driven Contrastive Mutual Learning for SAR Classification}

\author{
\IEEEauthorblockN{
 Pengfei Wang\textsuperscript{*}, 
 Hao Zheng\textsuperscript{*}, 
 Zhigang Hu,
 Aikun Xu,
 Meiguang Zheng\textsuperscript{\dag}\thanks{\textsuperscript{*} These authors contributed equally and \textsuperscript{\dag} is corresponding author.}, 
 and Liu Yang
 }
\IEEEauthorblockA{
School of Computer Science and Engineering, Central South University, Changsha, China\\
Email: \{234712162, zhenghao, zghu, aikunxu, zhengmeiguang, yangliu\}@csu.edu.cn
}
}

\maketitle

\begin{abstract}

Existing SAR image classification methods based on Contrastive Learning often rely on sample generation strategies designed for optical images, failing to capture the distinct semantic and physical characteristics of SAR data. To address this, we propose Physics-Driven Contrastive Mutual Learning for SAR Classification (PCM-SAR), which incorporates domain-specific physical insights to improve sample generation and feature extraction. PCM-SAR utilizes the gray-level co-occurrence matrix (GLCM) to simulate realistic noise patterns and applies semantic detection for unsupervised local sampling, ensuring generated samples accurately reflect SAR imaging properties. Additionally, a multi-level feature fusion mechanism based on mutual learning enables collaborative refinement of feature representations. Notably, PCM-SAR significantly enhances smaller models by refining SAR feature representations, compensating for their limited capacity. Experimental results show that PCM-SAR consistently outperforms SOTA methods across diverse datasets and SAR classification tasks.

\end{abstract}

\begin{figure*}
    \centering\includegraphics[width=0.98\linewidth]{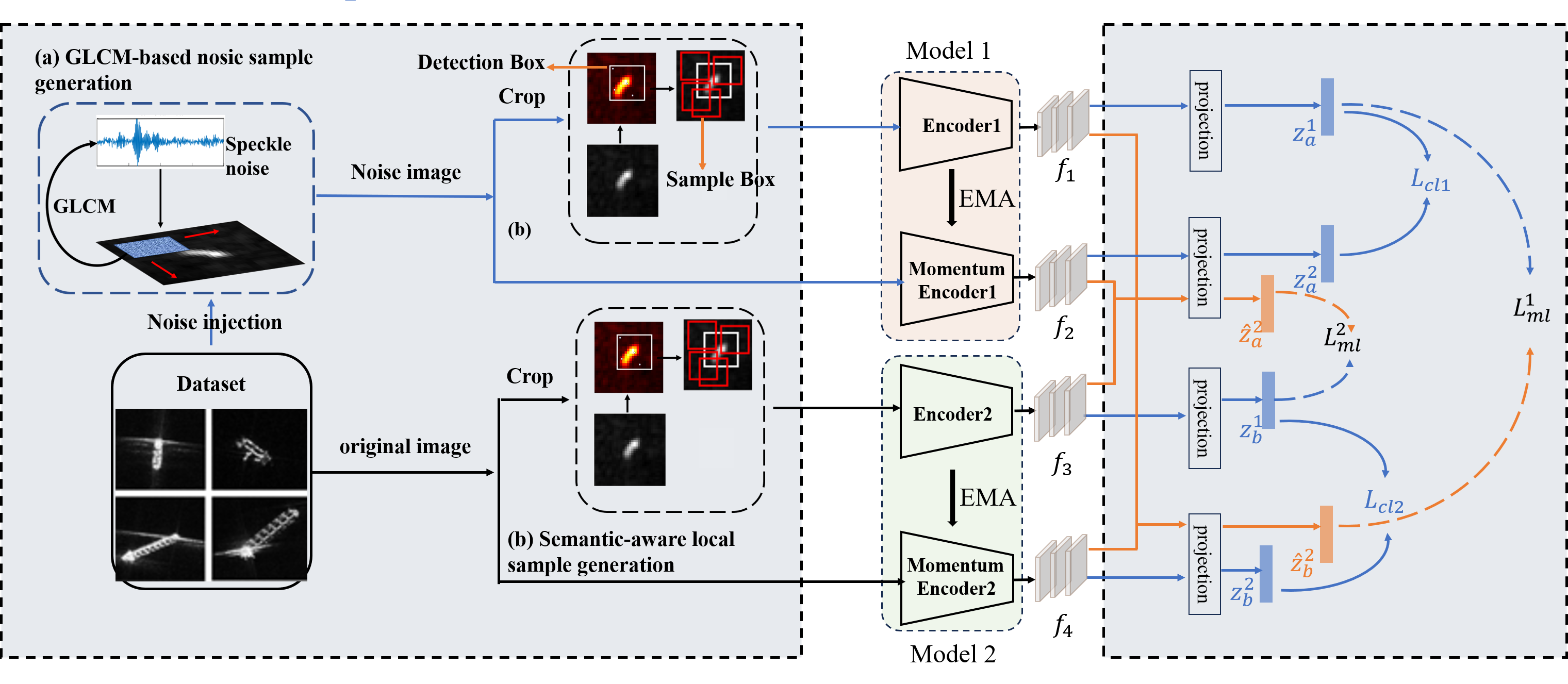}
     \caption{
     \small Overview of PCM-SAR. (a) is the GMCL nosie noise sample generation process, (b) is the sematic-aware sample generation process. Each model contains an encoder and a momentum encoder. The momentum encoder does not accept gradients but uses EMA(Exponential Moving Average) to update. The encoder accepts two types of losses: contrastive learning loss $L_{sm}$ and mutual learning loss $L_{ml}$. 
     }
    \label{fig:enter-label1}
\end{figure*}

\section{Introduction}

Synthetic Aperture Radar (SAR) has become an indispensable tool in remote sensing, providing all-weather, day-night imaging capabilities for critical applications such as marine monitoring, disaster management, and resource exploration \cite{SAR1}. However, automating SAR image classification faces significant challenges due to the scarcity of labeled data \cite{SAR2,SAR3,lin2024coarse}. Acquiring high-quality labels for SAR images is not only labor-intensive but also expensive, making supervised learning approaches less feasible for large-scale applications.

In recent years, Contrastive Learning (CL) has emerged as a promising unsupervised learning technique to alleviate the reliance on labeled data by training networks through positive and negative sample pairs to learn robust feature representations \cite{cl1}. Although CL has achieved success in optical image classification, its adaptation to SAR imagery is inherently challenging. Existing CL methods often employ sample generation strategies, such as introducing random Gaussian noise \cite{yang2022coarse,pei2023self} or random cropping \cite{kuang2024polarimetry,jiang2023adaptive,cao2023robust}, that fail to capture SAR’s unique semantic and physical characteristics. Unlike optical images, SAR data exhibits sparse and highly concentrated semantic information along with complex physical imaging properties, such as surface texture and speckle noise . These differences often lead to suboptimal sample representations and hinder the performance of CL-based SAR classification methods.

Traditional feature extraction methods, such as the gray-level co-occurrence matrix (GLCM) \cite{rangaiah2024histopathology,balling2023textural,ozturk2018application}, are designed to capture texture features and have proven effective in image classification tasks. However, their reliance on fixed scales and predefined patterns restricts their ability to adapt to the diverse and multi-scale textures inherent in SAR imagery. Furthermore, SAR data are characterized by physical structures and highly localized semantic information, which GLCM cannot adequately capture due to its lack of semantic awareness. This limitation prevents GLCM from interpreting global contexts or modeling the intricate imaging characteristics of SAR, such as speckle noise \cite{zhang2023blind,xu2024transedge}, spatial heterogeneity \cite{wang2023automatic}, and localized semantic regions \cite{shang2020semantic}, reducing its relevance in modern deep learning-based SAR frameworks.

To address these limitations, we propose Physics-Driven Contrastive Mutual Learning for SAR Classification (PCM-SAR), a novel framework that integrates domain-specific physical insights and semantic awareness into the sample generation process and feature extraction. PCM-SAR utilizes GLCM to simulate realistic noise patterns and employs a semantic-driven local sampling method to generate positive and negative sample pairs that better represent SAR’s unique imaging properties. This ensures that the generated samples reflect critical physical and semantic characteristics, enhancing the model’s ability to handle the heterogeneity of SAR data. Additionally, PCM-SAR incorporates a multi-level feature fusion mechanism based on mutual learning, enabling models to collaboratively refine their feature representations. This collaborative approach reduces reliance on individual positive-negative sample pairs while benefiting smaller models, which gain refined feature representations despite their limited capacity. Experimental results demonstrate that PCM-SAR consistently outperforms state-of-the-art methods across diverse datasets, achieving substantial improvements in SAR classification accuracy. In summary, the contribution of our method PCM-SAR is summarized as follows:
\begin{itemize}
\item We integrate SAR-specific physical insights into CL by utilizing GLCM to simulate realistic noise patterns and applying semantic-driven local sampling to generate positive and negative sample pairs. 
\item We propose a multi-level feature fusion mechanism based on ML, enabling models to collaboratively refine feature representations. This design not only enhances the extraction of SAR-specific features but also reduces reliance on individual positive-negative sample pairs.
\item we benefits smaller models by providing more refined feature representations, effectively compensating for their limited capacity. This improvement enables superior performance in SAR classification tasks across diverse datasets and model scales.
\end{itemize}
\label{sec:intro}

\section{METHODOLOGY}

\subsection{ 
Sample Generation Driven by Physical Mechanism
}
\label{sec:format}
\subsubsection{Noise Sample Generation(NSG) based on GLCM} 
The speckle noise in SAR images follows a multiplicative noise model \cite{xiang2023progressive, xiang2022optical}, where the the observed noisy image $\tilde{x}$ and the original image $x$ as follows:

\begin{equation}
\tilde{x}= x\cdot v_a\cdot e^{v_p},
\end{equation}
\noindent
where ${v_a}$ is the speckle amplitude noise, modeled as $v_{a}\sim {\mathcal{N} } \left ( 1, \sigma^{2}  \right ) $.
The intensity of speckle noise varies based on terrain features such as surface roughness, vegetation coverage, ocean currents, and wave activity, making noise modeling a complex task. 
To address this, we propose a speckle noise sample generation based on Gray-Level Co-occurrence Matrix (GLCM) \cite{shang2020semantic,james2021analysis}.
GLCM is a widely used texture analysis technique that extracts key features—such as Contrast (C), Entropy (E), and Homogeneity (H)—which represent the complexity and roughness of image blocks. These features allow us to account for terrain-induced noise variations in SAR images.
\begin{figure}[htpb]
    \centering
    \includegraphics[width=0.45\textwidth]{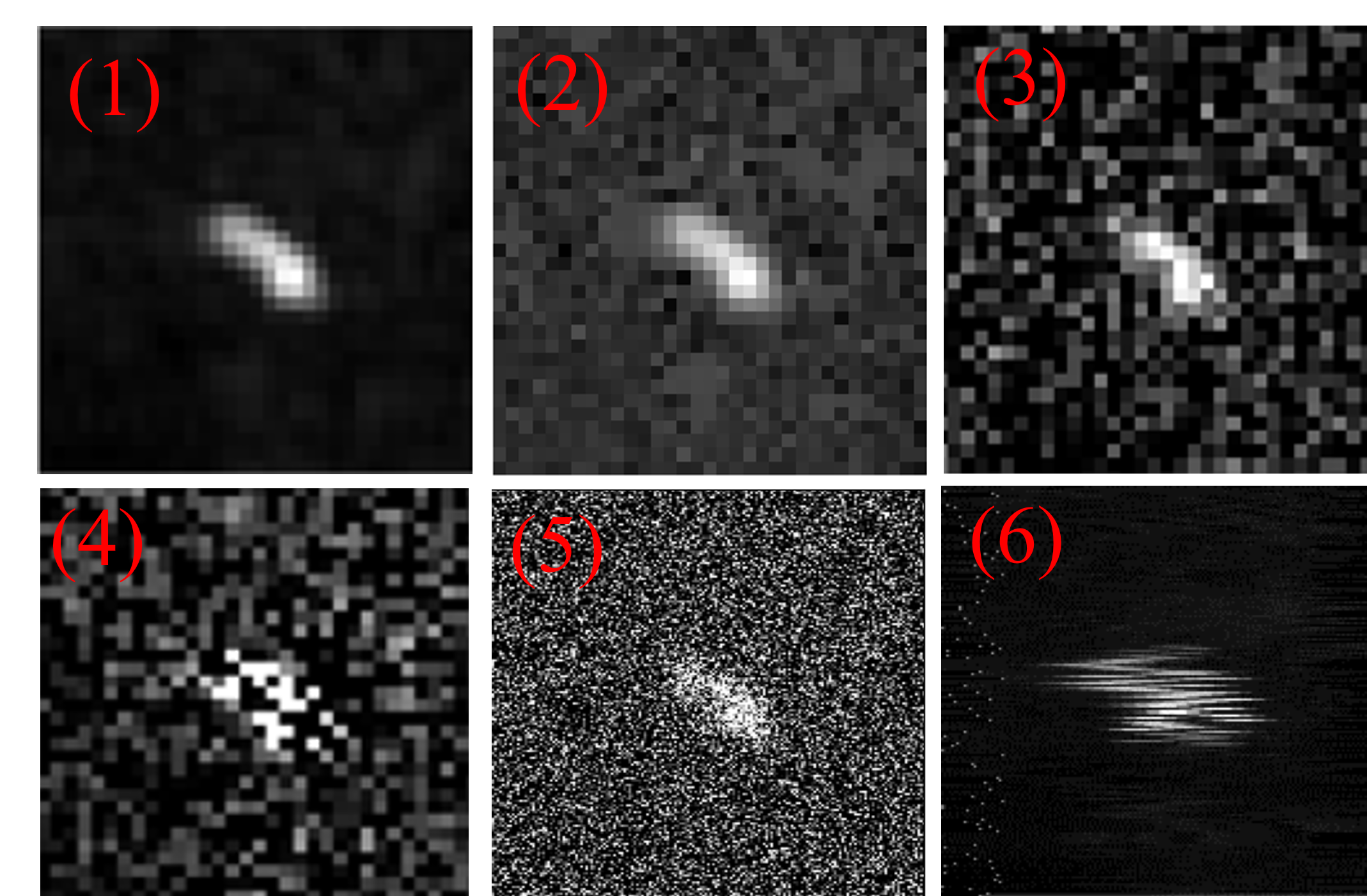}
    \caption{\small The images represent a comparison of five noise processing effects: (1) is original image, (2) is GLCM-based speckle noise, (3) is regular speckle noise, (4) is Gaussian noise(Commonly used in optical images), and (5) is scattering masking noise, (6) is time-shift noise. Among these, (3), (4), and (6) all significantly disrupt the semantic information of the original image, (5) will cover up the real texture and edge features in the image, which can negatively impact model training.}
    \label{fig:enter-label2}
\end{figure}

By using a variety of mathematical transformations to smooth the terrain and complexity effects, the final adjustment function $F$ is as follows:
\begin{equation}
    F=\alpha \cdot log(1+\beta _{1} \cdot C)+\gamma \cdot \sqrt{\beta _{2}\cdot E} +\delta \cdot exp(\beta _{3}\cdot H),
\end{equation}
where $C$, $E$, and $H$ are the GLCM-derived features, and $\alpha$, $\gamma$, $\delta$, and $\beta_1$, $\beta_2$, $\beta_3$ are tunable parameters. The variance of the speckle noise amplitude is adjusted using $\sigma^{\prime }  =\sigma _{0}\cdot F$, leading to an adjusted amplitude distribution $v_{a}^{\prime }\sim \mathcal{N} \left ( 1, (\sigma^{\prime })^{2} \right )$. The modified speckle noise intensity is then injected into the original image block to generate noise samples for contrastive learning tasks:
\begin{equation}
\tilde{x} = x\cdot v_{a}^{\prime } \cdot e^{v_p},
\end{equation}
where $v_a{\prime}$ is the adjusted noise amplitude based on $F$. This process produces a noise sample $\tilde{x}$ that closely approximates real-world SAR noise conditions.

\subsubsection{Semantic-aware Sample Generation(SSG)}
A single sample generation method may result in overfitting or underfitting of the model on some specific features, and building a high-quality sample space requires combining multiple sample generation schemes. 
Considering semantic information of SAR image is less and concentrated, we design an unsupervised local sampling scheme based on semantic perception.

\noindent
\textbf{Semantic-aware detection.}
For a SAR image $x$, $H(x)$ is heat map, bring the heat threshold $T$ into the activation function $l$ to generate a binary mask $M$:
\begin{equation}
    M=l(H(x),T)=\begin{cases}1,\quad if \quad H(x) \ge  T
 \\
0,\quad if \quad H(x) <  T,
\end{cases}
\end{equation}
the region in $x$ with brightness higher than $T$ is filtered out by a binary mask $M$, and the detection box $B$ is generated:
\begin{equation}
    B=bounding box\left ( x,M \right ),
\end{equation}
the resulting $B$ is a quadruple containing the coordinates of the four vertices of the detection box.

\noindent
\textbf{Decentralization sampling.}
Using the Poisson disk sampling algorithm to generate sample points $p$ in the detection frame: 
\begin{equation}
\begin{split}
p=&\left \{p_{i}\,\mid p_{i}\in B,\forall_{ j\ne i }  ,\left \| p_{i}-p_{j} \right \|> r ,\right.\\
&\left.p_{i}=random(p_{k}+d_{i}\cdot \beta ,\beta \sim \beta(\mathbb{S}^{1} ) )\right \}, 
\end{split}
\end{equation}
where $p_{k}$ are valid sampling points from the previous iteration, $r$ is the minimum distance constraint ensuring even distribution, $\beta$ is a random vector sampled from the specific distribution, and $d_i$ is a direction vector for randomization. The generated points $p$ serve as the central coordinates for sampling boxes, ensuring that the sampling process is both decentralized and comprehensive. This approach generates a diverse set of localized samples that reflect the semantic and structural characteristics of SAR images.

By combining NSG and SSG, the sample space becomes more diverse and representative, encompassing various noise and local semantic conditions encountered in real-world SAR data. This enriched sample space improves the model’s sensitivity to local details and enhances its ability to generalize across complex SAR image features.


\subsection{ Mutil-level Feature Fusion(FF) Based on Mutual Learning}

\subsubsection{Feature Fusion Process}

\begin{figure}[t]
    \centerline{\includegraphics[width=1\columnwidth]{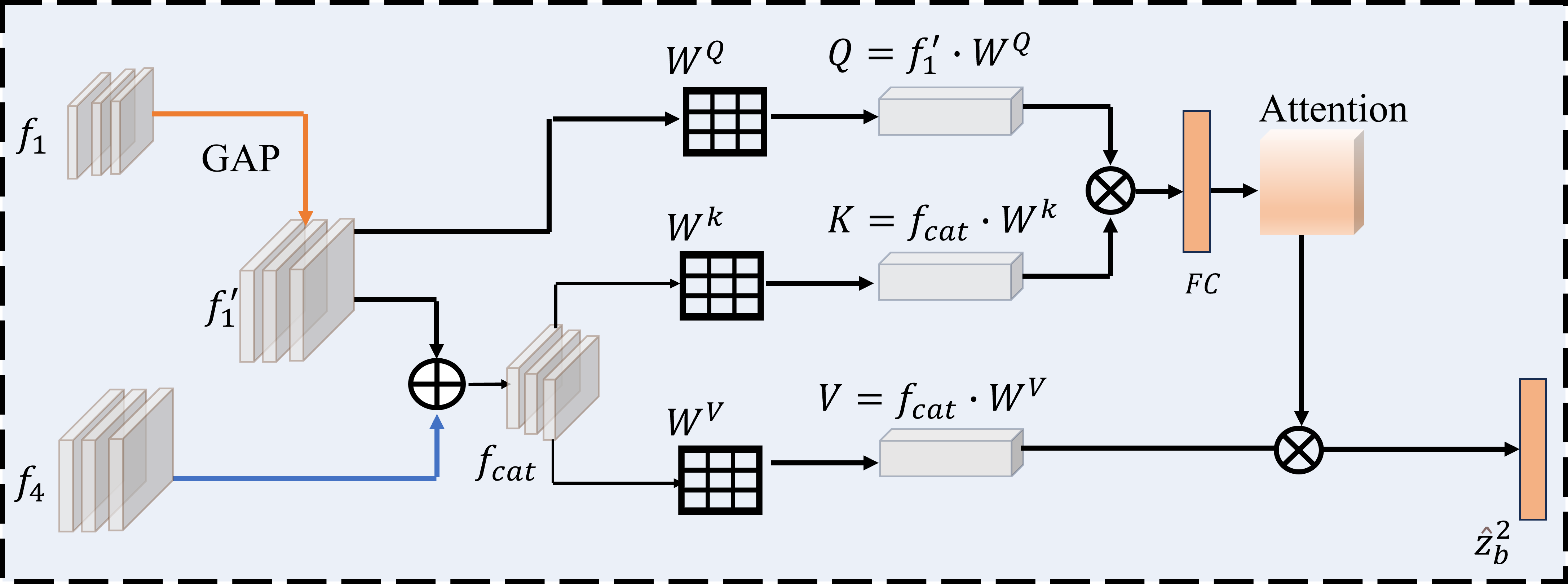}}
    \caption{The feature fusion process inside the projection head.}
    \label{fig:enter-label3}
\end{figure}

As show in Fig.\ref{fig:enter-label3}, the feature fusion mechanism  integrates local and global features to enhance the model’s representation capability. Specifically, the projection head combines local features $f_1$ and global features $f_4$, where $f_{1}^{\prime }$ is derived from $f_1$ via global average pooling. These are concatenated to form $f_{cat}$, a comprehensive feature vector representing both local and contextual information. To refine feature interactions, the projection head uses learnable parameter matrices $W^Q$, $W^K$, and $W^V$ to compute the query (Q), key (K), and value (V) vectors, respectively. An attention mechanism calculates the output as:
\begin{equation}
\hat{z}_{b 2}=\operatorname{softmax}\left(\frac{\left(W^{Q} f_{1}\right)\left(W^{K} f_{4}\right)^{T}}{\sqrt{d_{k}}}\right) \cdot\left(W^{V} \cdot f_{cat})\right),
\end{equation}
where $d_{k}$ is the dimension of the key vector. 
Through this feature fusion process, the weights of the generated vectors are redistributed, enabling the model to have stronger global information capture capabilities and further enhancing feature extraction capabilities, and better integrates the physical mechanisms of SAR images into the sample space. By integrating local and global features, noise and raw features, the model can more accurately simulate real SAR images, thus better reflecting the essential characteristics in the sample space.

\subsubsection{Loss Calculation}
The loss calculation process involves two primary components: CL Loss and ML Loss. In Fig.\ref{fig:enter-label3}, representations $f_1$ and $f_2$ are processed through the projection head to produce embeddings $z_{a1}$ and $z_{a2}$, where $z_{a1}$, $z_{a2} \in \mathbb{R}^k$ and $k$ is the output dimension. The embeddings are normalized using the softmax function:

\begin{equation}
    z_{a1}^{i}=\frac{exp\left ( z_{a1}^{i}/t \right ) }{ {\textstyle \sum_{k=1}^{K}exp\left ( z_{a1}^{k}/t \right ) } },  
\end{equation}
where \(t\) is a temperature scaling parameter. The contrastive loss for each embedding pair is calculated as:
\begin{equation}
    \mathcal{L}_{cl1}=- \frac{1}{2} \left ( z_{a1}log\left (  z_{a2}\right )  + z_{a2}log\left (  z_{a1}\right )\right ),
\end{equation}
\begin{equation}
    \mathcal{L}_{cl2}=- \frac{1}{2} \left ( z_{b1}log\left (  z_{b2}\right )  + z_{b2}log\left (  z_{b1}\right )\right ),
\end{equation}
Similarly, for the mutual learning process, the enhanced vector $\hat{z}$, which integrates shared semantic information, is incorporated to compute the ML loss:
\begin{equation}
    \mathcal{L}_{ml}^{1} =-\frac{1}{2}(z_{a1}log(\hat{z} _{b2} ) +\hat{z} _{b2}log(z_{a1} )),
\end{equation}
\begin{equation}
    \mathcal{L}_{ml}^{2} =-\frac{1}{2}(z_{a2}log(\hat{z} _{b1} ) +\hat{z} _{b1}log(z_{a2} )), 
\end{equation}
where \(\hat{z}\) incorporates shared semantic information between global and local features. 

The total loss for each model is the weighted sum of CL loss and ML loss:

\begin{equation}
    \left \{ \begin {array}{l} {\mathcal {L}_1} = \lambda _{1}{\mathcal {L}_{cl1}} + \lambda _{2} {\mathcal L_{ml}^{1}}\\ {\mathcal {L}_2} = \lambda _{1}{\mathcal {L}_{cl2}} + \lambda _{2}{\mathcal L_{ml}^{2}} \end {array} \right . \label {equ:l_all} ,
\end{equation}
\begin{table*}[t!] 
\centering
\renewcommand{\arraystretch}{1.3}
\caption{The test accuracy (\%) comparisons of proposed PCM-SAR with traditional GLCM methods and ordinary speckle noise sample generation schemes on OpenSARShip and FUSARShip datasets.}
\label{table:traditional}
\begin{tabular}{lcccccc}
\toprule
\textbf{Methods}      & Baseline & Gaussian Noise \cite{yang2022coarse} & Texture Extraction \cite{balling2023textural} & Speckle Noise \cite{kuang2024polarimetry} & \textbf{PCM-SAR(Ours)} & \textbf{Improvement(\%)} \ \\ 
\midrule
OpenSARShip & 69.43 & 69.87  & 70.07 & \underline{71.22} & \textbf{72.32} & 1.10 $\uparrow$ \\
FUSARShip & 75.31  & 75.18 & 76.24  & \underline{77.12} & \textbf{88.64} & 11.52 $\uparrow$ \\
\bottomrule
\end{tabular}
\end{table*}
\noindent

\begin{table}[ht]
\centering
\caption{The test accuracy (\%) comparisons of proposed PCM-SAR with SOTA contrastive learning methods.}
\label{table:SOTA}
\renewcommand{\arraystretch}{1.2}
\resizebox{\linewidth}{!}{
\begin{tabular}{l|cc|cc}
\toprule
\multirow{2}{*}{\textbf{Methods}} & \multicolumn{2}{c}{\textbf{FUSAR-Ship}} & \multicolumn{2}{|c}{\textbf{OpenSARShip}} \\ 
\cmidrule(lr){2-3} \cmidrule(lr){4-5}
& \textbf{K-NN} & \textbf{LP} & \textbf{K-NN} & \textbf{LP} \\ 
\midrule
Baseline & 77.45 $\pm$ 2.56 & 75.32 $\pm$ 1.45 & 68.06 $\pm$ 1.12 & 69.45 $\pm$ 0.92 \\
SimCLR\cite{cao2023robust} & 80.77 $\pm$ 0.33 & 78.23 $\pm$ 0.53 & 69.34 $\pm$ 2.06 & 68.99 $\pm$ 1.51 \\
Disco\cite{gao2022disco} & 79.09 $\pm$ 0.96 & 81.45 $\pm$ 0.42 & 68.18 $\pm$ 1.88 & 68.53 $\pm$ 1.14 \\
PCL \cite{kuang2024polarimetry} & 86.26 $\pm$ 2.11 & 86.79 $\pm$ 0.45 & 69.94 $\pm$ 1.88 & 70.21 $\pm$ 1.95 \\
CSFL\cite{yang2022coarse} & 87.09 $\pm$ 0.48 & 86.24 $\pm$ 0.74 & 70.36 $\pm$ 2.20 & 70.18 $\pm$ 1.67 \\
DCPN\cite{cl1} & \underline{87.94} $\pm$ 0.76 & 85.32 $\pm$ 0.40 & 70.33 $\pm$ 2.27 & \underline{71.17} $\pm$ 0.81 \\
SAIM \cite{song2024semantic} & 85.23 $\pm$ 0.45 & \underline{86.87} $\pm$ 0.77 & \underline{71.03} $\pm$ 1.23 & 70.99 $\pm$ 0.64 \\
\midrule
\textbf{PCM-SAR(Ours)} & \textbf{88.23 $\pm$ 0.47} & \textbf{88.48 $\pm$ 0.29} & \textbf{72.10 $\pm$ 1.63} & \textbf{72.48 $\pm$ 2.04} \\ 
\midrule
\textbf{Improvement(\%)} & 0.29 $\uparrow$ & \textbf{1.61} $\uparrow$ & 1.07 $\uparrow$ & 1.31 $\uparrow$ \\ 
\bottomrule
\end{tabular}
}
\end{table}
\noindent
where $\lambda_1 \in [0, 1]$ and $\lambda_2 \in [0, 1]$ are tunable hyperparameters that control the balance between the CL loss and ML loss. $\lambda_1$ emphasizes the alignment of positive and negative sample pairs, while $\lambda_2$ focuses on refining features through multi-level mutual learning.

\section{EXPERIMENT AND RESULT ANALYSIS}
The experiments are conducted on the OpenSARShip \cite{huang2017openSARship} and FUSAR-Ship \cite{hou2020FUSAR} datasets. Following pre-training, Linear Probing (LP) and K-Nearest Neighbor (K-NN) are performed to evaluate the model’s effectiveness. The results are reported as “mean$\pm$std” accuracy. For all experiments, ResNet series models \cite{he2016deep} are employed as the backbone network.

\subsection{Compared with traditional GLCM methods and ordinary speckle noise sample generation schemes}
As shown in the Table \ref{table:traditional}, the sample generation method using Gaussian random noise yields the worst performance, as the noise in SAR is predominantly speckle noise. The second is the sample generation scheme using texture extraction via GLCM and the ordinary speckle noise generation scheme. The former simply performs ordinary texture enhancement, and the latter does not make targeted improvements based on its physical characteristics in the process of generating speckle noise. After using the GLCM matrix to obtain the texture information of the image, our method designs an adjustment function for the speckle noise amplitude, integrating the
physical characteristics of the SAR image into the samples of contrastive learning, the final classification accuracy is improved by 2.89\% on OpenSARShip and 13.33\% on FUSARShip compared to the baseline.

\begin{table}[t]
\centering
\renewcommand{\arraystretch}{1.2} 
\setlength{\tabcolsep}{2pt} 
\caption{Impact of Different Weight Combinations of CL and ML Loss ($\lambda_1, \lambda_2$) on Test Accuracy for Various ResNet Backbones }
\label{parament}
\begin{tabular}{{c|c|c|c}}
\toprule
Backbone &   \(\lambda_{1}\) = 1.0, \(\lambda_{2}\) = 0  &   \(\lambda_{1}\) = 1.0, \(\lambda_{2}\) = 0.5  &   \(\lambda_{1}\) =1.0, \(\lambda_{2}\) = 1.0  \\
\midrule
Resnet50&74.25&74.56&74.58\\
Resnet34&73.34&75.11&76.11\\
Resnet18&72.25&75.16&76.21\\
\bottomrule
\end{tabular}
\end{table}

\subsection{Comparison with the SOTA CL Methods}
In order to further prove the advantage of our PCM-SAR method, we compared with many recent CL methods on SAR images. As show in Table \ref{table:SOTA}, we have observed that schemes used on general-purpose optical datasets often do not yield better classification results on SAR image datasets, for example, compared with Disco and SimCLR  for general optical datasets, PCM-SAR achieved up to 9.14\% and 7.46\% improvement on the FUSARShip dataset (K-NN), and 3.92\% and 3.51\% improvement on the OpenASARShip dataset (K-NN), because these methods often adopt sampling methods for optical data, they lack attention to the physical characteristics and semantic information of SAR images. Compared with the PCL method using ordinary speckle noise samples, our method improves 1.97\% (K-NN) and 1.67\% (LP) on the FUSARShip dataset, because the ordinary speckle noise model assumes that the variance of its amplitude follows a normal distribution without considering its changes from the perspective of its true physical properties. Compared with CSFL, DCPN that use random cropping to generate local samples, our method improves by 1.14\% and 0.29\% (K-NN) on the FUSARShip dataset and by 2.30\% and 1.31\% (LP) on the OpenSARShip dataset. This is because random sample generation is difficult to capture the pivotal semantic information of SAR images and is prone to generate poor quality comparison samples.

\subsection{Effect of hyper-parameters $\lambda _{1}$  and $\lambda _{2}$ }
As shown in Table \ref{parament}, we found that with the increase of ML loss weight $\lambda _{2}$, the performance of all models was improved. Among them, in the large parameter scale model Resnet50, as the ML loss weight $\lambda _{2}$ increases from 0 to 1, the accuracy is improved by 0.31\% and 0.33\%, respectively. On the contrary, for Resnet34 and Resnet18 with smaller parameter scale, as the ML loss weight increases, the improvement in accuracy is significantly increased. We find when $\lambda _{2}$ increases from 0 to 1, the accuracy of Resnet34 is improved by 0.68\% and 2.78\%, respectively, and the accuracy of Resnet18 is improved by 2.91\% and 3.96\%, respectively. 

\begin{table*}[t]
\centering
\renewcommand{\arraystretch}{1.2}
\caption{Ablation Study of the PCM-SAR Framework on OpenSARShip Dataset with Baseline Models and Downstream Tasks K-NN and LP}
\label{table:Ablation}
\resizebox{\textwidth}{!}{%
\begin{tabular}{@{}cc|cc|cc|cc|cc|cc@{}}
\toprule
\multicolumn{2}{c|}{\textbf{Backbone}} & \multicolumn{2}{c|}{\textbf{Baseline}} & \multicolumn{2}{c|}{\textbf{Without NSG}} & \multicolumn{2}{c|}{\textbf{Without SSG}} & \multicolumn{2}{c|}{\textbf{Without FF}} & \multicolumn{2}{c}{\textbf{PCM-SAR (Ours)}} \\ 
\cmidrule(r){1-2} \cmidrule(r){3-4} \cmidrule(r){5-6} \cmidrule(r){7-8} \cmidrule(r){9-10} \cmidrule(r){11-12}
\textbf{Basenet 1} & \textbf{Basenet 2} & \textbf{K-NN} & \textbf{LP} & \textbf{K-NN} & \textbf{LP} & \textbf{K-NN} & \textbf{LP} & \textbf{K-NN} & \textbf{LP} & \textbf{K-NN} & \textbf{LP} \\ 
\midrule
\multirow{6}{*}{Resnet50} & \multirow{2}{*}{Resnet50} & 68.06$\pm$1.12 & 68.25$\pm$0.92 & 70.35$\pm$0.88 & 71.46$\pm$1.12 & 70.59$\pm$0.25 & 69.54$\pm$5.77 & 69.24$\pm$2.07 & 68.48$\pm$3.52 & \textbf{71.10$\pm$1.63} & \textbf{72.67$\pm$2.04} \\
         &           &  ---               &  ---              & (+2.29)       & (+3.40)       & (+2.53)       & (+1.48)       & (+1.18)       & (+0.48)       & (+3.04)       & (+4.42)       \\ 
 & \multirow{2}{*}{Resnet34} & 70.81$\pm$0.12 & 71.06$\pm$1.33 & 72.65$\pm$0.89 & 71.99$\pm$3.87 & 71.23$\pm$3.61 & 71.35$\pm$4.02 & 71.39$\pm$3.58 & 71.55$\pm$1.89 & \textbf{74.90$\pm$0.62} & \textbf{74.06$\pm$1.39} \\
         &           &  ---               &  ---              & (+1.82)       & (+0.93)       & (+0.42)       & (+0.29)       & (+0.49)       & (+0.49)       & (+4.09)       & (+3.26)       \\ 
 & \multirow{2}{*}{Resnet18} & 70.56$\pm$1.19 & 69.06$\pm$1.55 & 72.09$\pm$0.45 & 71.61$\pm$3.87 & 70.89$\pm$2.51 & 69.88$\pm$2.77 & 70.82$\pm$1.84 & 69.62$\pm$2.73 & \textbf{73.26$\pm$1.88} & \textbf{73.75$\pm$2.32} \\
         &           &  ---               &  ---              & (+1.53)       & (+2.55)       & (+0.33)       & (+0.82)       & (+0.26)       & (+0.26)       & (+2.77)       & (+3.26)       \\ 
\midrule      
\multirow{4}{*}{Resnet34} & \multirow{2}{*}{Resnet34} & 69.99$\pm$2.16 & 67.81$\pm$2.82 & 71.32$\pm$2.03 & 71.28$\pm$2.93 & 71.89$\pm$1.29 & 70.62$\pm$3.16 & 70.03$\pm$3.08 & 68.72$\pm$1.12 & \textbf{73.26$\pm$2.42} & \textbf{73.04$\pm$1.86} \\
         &           & ---                & ---               & (+1.33)       & (+3.42)       & (+1.90)       & (+2.82)       & (+0.33)       & (+0.87)       & (+3.27)       & (+3.05)       \\ 
& \multirow{2}{*}{Resnet18} & 69.92$\pm$0.77 & 69.09$\pm$2.11 & 71.26$\pm$0.11 & 70.63$\pm$3.87 & 70.66$\pm$1.42 & 70.48$\pm$1.12 & 70.96$\pm$0.82 & 72.42$\pm$1.39 & \textbf{72.96$\pm$0.82} & \textbf{72.42$\pm$1.39} \\
         &           &  ---               & ---               & (+1.36)       & (+1.54)       & (+0.74)       & (+1.39)       & (+0.10)       & (+0.75)       & (+3.04)       & (+2.50)       \\ 
\midrule
\multirow{2}{*}{Resnet18} & \multirow{2}{*}{Resnet18} & 69.23$\pm$1.47 & 70.01$\pm$1.39 & 70.85$\pm$1.12 & 70.69$\pm$3.87 & 69.52$\pm$1.81 & 70.06$\pm$2.24 & 70.88$\pm$1.75 & 70.36$\pm$2.52 & \textbf{72.44$\pm$1.38} & \textbf{72.63$\pm$2.24} \\
         &           & ---                &  ---              & (+1.72)       & (+0.68)       & (+0.29)       & (+0.29)       & (+1.65)       & (+0.35)       & (+3.21)       & (+3.40)       \\ 

\bottomrule
\end{tabular}
}
\end{table*}

\begin{table}[t]
\centering
\renewcommand{\arraystretch}{1.2}
\setlength{\tabcolsep}{2pt}
\caption{The Performance of PCM-SAR with Different Backbones and Label Ratios (10\% and 30\% Labels) in Semi-Supervised Scenarios on FUSARShip dataset.}
\label{table:Semi}
\resizebox{\linewidth}{!}{
\begin{tabular}{cc|cc|cc|cc|cc}
\toprule
\multicolumn{2}{c|}{\textbf{Backbone}}  & \multicolumn{4}{c|}{\textbf{10\% Labels}} & \multicolumn{4}{c}{\textbf{30\% Labels}}\\ 
\cmidrule(lr){1-2} \cmidrule(lr){3-6} \cmidrule(lr){7-10}
\textbf{Basenet1} & \textbf{Basenet2}  & \multicolumn{2}{c|}{\textbf{Independent}} & \multicolumn{2}{c|}{\textbf{PCM-SAR}} & \multicolumn{2}{c|}{\textbf{Independent}} & \multicolumn{2}{c}{\textbf{PCM-SAR}} \\ 
\midrule
Resnet50                          & Resnet50        & 47.44           & 43.16           & \textbf{54.32}  & 46.21           & 69.01           & 66.89           & \textbf{69.17}  & 68.02           \\
Resnet34                          & Resnet34        & 49.31           & 57.05           & \textbf{59.28}  & 59.03           & 67.35           & 72.94           & 71.20           & \textbf{75.03}  \\
Resnet18                          & Resnet18        & 54.89           & 43.96           & \textbf{58.91}  & 43.88           & 70.72           & 67.63           & \textbf{71.80}  & 67.08           \\
Resnet50                          & Resnet34        & 48.95           & 43.66           & \textbf{49.67}  & 43.81           & 67.39           & 62.98           & \textbf{67.39}  & 69.05           \\
Resnet50                          & Resnet18        & 49.85           & 38.98           & \textbf{51.24}  & 41.88           & 67.37           & 56.09           & \textbf{67.33}  & 69.87           \\
Resnet34                          & Resnet18        & 35.89           & 35.08           & \textbf{39.25}  & 36.46           & 57.02           & 55.90           & \textbf{59.32}  & 56.31           \\ 
\bottomrule
\end{tabular}
}
\end{table}

\subsection{Ablation experiments}
As shown in Table \ref{table:Ablation}, our proposed PCM-SAR can improve the accuracy of almost all backbones. In the K-NN \cite{cheng2014k}, the combination of Resnet50 and Resnet34 has achieved the highest accuracy improvement, and the accuracy has increased by 4.09$\%$. In the linear detection \cite{kumar2022fine}, resnet-50 as the backbone achieved the best result, with an accuracy improvement of 4.42$\%$. Correspondingly, we found that models with larger parameter sizes do not necessarily perform better.

\subsection{Transfer to Semi-Supervised Learning Scenarios}
We established two semi-supervised learning scenarios to simulate SAR image classification in settings with limited labeled samples. As shown in Table \ref{table:Semi}, all models trained under the PCM-SAR framework achieved higher accuracy than those trained independently. More importantly, we found that the isomorphic combination of ResNet34 and ResNet18, which have smaller parameter scales, achieved the highest accuracies (75.03\% and 71.80\%) on the 30\% labeled dataset. Compared to the ResNet50 model with a larger parameter scale under the same conditions, the performance improved by 2.63\% and 7.01\%, respectively. In the training setting with 10\% labeled data, the ResNet18 and ResNet34 models again demonstrated better performance (59.28\% and 58.91\%), improving by 4.96\% and 4.59\%, respectively, compared to the ResNet50 model. These results show that our method alleviates the burden of extracting and understanding complex features for small parameter scale models by refining SAR image representation, offering superior performance and improvement rates compared to baseline and large parameter scale models in most environmental settings.

\begin{table}[t]
\centering
\renewcommand{\arraystretch}{1.2} 
\setlength{\tabcolsep}{2pt} 
\caption{Performance Comparison of PCM-SAR and Other Methods on Detection and Segmentation Tasks with Different Backbones on FUSARShip. Mask R-CNN~\cite{hedoll} is adopted as the detector.}
\label{table:detection}
\begin{tabular}{@{}lcc|ccc|ccc@{}}
\toprule
\multirow{2}{*}{\textbf{Method}}    & \multirow{2}{*}{\textbf{Basenet1}}    & \multirow{2}{*}{\textbf{Basenet2}}    & \multicolumn{3}{c|}{\textbf{Detection}}                  & \multicolumn{3}{c}{\textbf{Segmentation}}               \\ 
\cmidrule(lr){4-6} \cmidrule(lr){7-9}
                   &                  &                  & $AP^b$    & $AP^b_{50}$ & $AP^b_{75}$ & $AP^s$    & $AP^s_{50}$ & $AP^s_{75}$ \\ 
\midrule
Simclr [19]        & Resnet18         & Resnet18         & 38.45     & 57.28       & 42.08       & 33.44     & 53.39       & 34.80       \\
Disco [9]          & Resnet18         & Resnet18         & 38.94     & 57.92       & 41.81       & 35.03     & 56.16       & 35.99       \\
PCL [3]            & Resnet18         & Resnet18         & 40.07     & 59.01       & 42.92       & 34.88     & 56.63       & 36.45       \\
PCM-SAR            & Resnet18         & Resnet18         & \textbf{41.73} & \textbf{59.97} & \textbf{61.22} & \textbf{35.09} & \textbf{58.59} & \textbf{38.12} \\ 
\midrule
Simclr [19]        & Resnet50         & Resnet50         & 38.05     & 56.93       & 41.42       & 33.68     & 54.04       & 35.77       \\
Disco [9]          & Resnet50         & Resnet50         & 39.22     & 58.29       & 42.04       & 34.38     & 55.66       & 36.12       \\
PCL [3]            & Resnet50         & Resnet50         & 39.49     & 59.08       & 42.36       & 34.68     & 56.98       & 36.59       \\
PCM-SAR            & Resnet50         & Resnet50         & \textbf{40.33} & \textbf{59.36} & \textbf{43.57} & \textbf{34.85} & \textbf{57.99} & \textbf{36.95} \\ 
\bottomrule
\end{tabular}
\end{table}

\subsection{Transfer to Detection and Segmentation Tasks}
As shown in Table \ref{table:detection}, by different IOU settings, PCM-SAR achieved the highest accuracy in both object detection and semantic segmentation. Among them, The object detection task, using the ResNet18 model as the backbone with $AP^b_{75}$, and the semantic segmentation task, using the ResNet18 model as the backbone with $AP^s_{50}$, achieved the highest accuracies of 61.22\% and 58.59\%, respectively. These results were 19.14\% and 5.20\% higher than the traditional Simclr method, and also surpassed the ResNet50 model under the PCM-SAR by 2.95\% and 3.95\%. Compared with the common speckle noise sample generation method, our method improves the target detection task by 19.76\% by $AP^b_{75}$, and the semantic segmentation task by 1.96\% by $AP^s_{50}$. This further reflects the advantage of PCM-SAR in improving the performance of smaller parameter scale models, and proves that PCM-SAR can improve the positive and negative samples of contrastive learning by integrating physical insights and semantic information capture, and has good generalization ability on dense prediction tasks.

\section{CONCLUSION}
We propose PCM-SAR which is a specially designed for SAR image classification by leveraging the physical mechanisms of SAR. PCM-SAR employs a GLCM-based approach to generate realistic noise samples and incorporates semantic-aware localization to obtain diverse local samples with higher coverage. This enhances the quality and diversity of positive and negative samples in contrastive learning. Additionally, the introduction of a mutual learning framework with a multi-level feature fusion mechanism enables models to share and refine feature representations, significantly improving feature extraction and alignment with the demands of SAR-specific data. Experimental results demonstrate that PCM-SAR consistently outperforms existing contrastive learning methods. 

\section{Acknowledgements}
We appreciate constructive feedback from anonymous reviewers and meta-reviewers. This work was supported by the National Natural Science Foundation of China (62172442, 62172451), China Scholarship Council, and High Performance Computing Center of Central South University.

\bibliographystyle{IEEEbib}
\bibliography{pcm_sar}

\color{red}

\end{document}